\pdfoutput=1

\documentclass[11pt]{article}

\usepackage{acl}

\usepackage{times}
\usepackage{latexsym}
\usepackage{amsmath,amsfonts,amsthm,bm} 
\usepackage{graphicx}
\usepackage[T1]{fontenc}

\usepackage[utf8]{inputenc}

\usepackage{microtype}

\title{Latent Aspect Detection from Online Unsolicited Customer Reviews}

\author{Mohammad Forouhesh \\
  Amirkabir University of Technology, Iran \\
  \texttt{mforouhesh@aut.ac.ir} \\\And
  Arash Mansouri \\
  Press'nXPress, Canada \\
  \texttt{mansouri@pressnxpress.com} \\\AND
  Hossein Fani \\
  University of Windsor, Canada \\
  \texttt{hfani@uwindsor.ca} \\}

\begin{document}
    \maketitle
\begin{abstract}
Within the context of review analytics, aspects are the features of products and services at which customers target their opinions and sentiments. Aspect detection helps product owners and service providers to identify shortcomings and prioritize customers' needs, and hence, maintain revenues and mitigate customer churn. Existing methods focus on detecting the surface form of an aspect by training supervised learning methods that fall short when aspects are latent in reviews. In this paper, we propose an unsupervised method to extract latent occurrences of aspects. Specifically, we assume that a customer undergoes a two-stage hypothetical generative process when writing a review: (1) \textit{deciding} on an aspect amongst the set of aspects available for the product or service, and (2) writing the opinion words that are more interrelated to the chosen aspect from the set of all words available in a language. We employ latent Dirichlet allocation to learn the latent aspects distributions for generating the reviews. Experimental results on benchmark datasets show that our proposed method is able to improve the state of the art when the aspects are latent with no surface form in reviews. 
\end{abstract}

\section{Introduction}
The key characteristic of customers' opinions is what they target in order to convey a sentiment, referred to as aspect (variously known as features, properties, targets, attributes, or components). For example, in the review \textit{`advertise free wifi but charge you for that,'} the aspect is \textit{`advertisement'} toward which an opinion \textit{`misinformation'} is expressed with \textit{`negative'} sentiment. Detection of aspects is a crucial step in the analysis of customers' reviews for products or services in e-commerce and social platforms as it helps product owners and service providers to find shortcomings and prioritize the improvements according to customers needs, and hence, maintaining revenues and decrease customer churn~\cite{Napitu_2017}. As such, aspect detection along with customers' opinion mining and sentiment analysis has long gained much attention~\cite{Cortis_2021}. 

\begin{table}
\centering
\begin{tabular}{@{}l}
\hline
\textbf{Reviews: [{\color{gray}latent}] vs. {\color{blue}\underline{explicit}} aspects} \\
\hline
\textit{-were given table far from river }[{\color{gray}management}] \\
\textit{-the girl} [{\color{gray}staff}] \textit{at the front desk was really nice} \\
\textit{-they {\color{blue}\underline{advertise}} free wifi but charge you for that}\\ 
\textit{-{\color{blue}\underline{menu}} is extensive and a {\color{blue}\underline{bar}} with live music} \\
\hline
\end{tabular}
\caption{Reviews for restaurants from Google reviews.
}\label{examples-hidden-explicit}
\end{table}

Existing methods to detect aspects in reviews and their respective sentiment with the opinions fall into three categories based on the level of human supervision: \textit{i}) rule-based, \textit{ii}) supervised, and \textit{iii}) unsupervised methods. Rule-based methods use the co-occurrence association rule mining approach to match aspects with explicit surface forms (words) \cite{Hai_2011, Zeng_2013}. However, such methods are not scalable to an overwhelming number of combinations in reviews. Supervised methods employ supervised machine learning techniques on labeled datasets where aspects' surface forms are manually annotated in reviews \cite{Peng_2020, Zhang_2020}. To eschew reliance on labeled data whose curation is time-consuming, prone to human biases, and has high manpower cost, unsupervised methods have been adopted \cite{He_2017, Zong_2021}. Proposed unsupervised methods, however, assume that aspects are expressed explicitly in customers' reviews and forego the latent occurrences of aspects in reviews. Indeed, when writing an online unsolicited review, a customer may write about her opinion and overall rating while overlooking mentioning the aspects' surface form for a product or service due to their trivialities or being common knowledge. In Table~\ref{examples-hidden-explicit}, some samples from Google reviews have been shown. For instance, the restaurant review \textit{`we were given table far from river view'} entails one or several aspects \textit{`management'} or \textit{`staff'} that are latent and not explicitly mentioned (do not have surface form) due to the background knowledge on restaurants common services. Indeed, 35\% of reviews on restaurants and electronics include latent aspects \cite{Cai_2021,Xu_2015}. The need for latent aspects detection is even more pressing in online social review platforms where reviews are short, noisy, informal, and mostly rely on social background knowledge and context.

To this end, in this work, we consider the task of \textit{latent} aspect detection. Specifically, given a collection of reviews with no human supervision (no labels for aspects, opinion, nor sentiment and whether of their explicit or latent appearances), we aim at finding latent (hidden) aspects in customer reviews. Although \citet{Marrese_Taylor_2013, Poria_2014, Chen_2016, Wan_2020} have attempted to solve the latent aspect detection problem, they all focus on supervised methods on labeled datasets provided, e.g., SemEval-2016 \cite{Pontiki_2016}. In this work, unlike the existing studies, we have no direct access to the aspects' annotations. 

Inspired by LDA \cite{Blei_2003}, we propose to model aspects as latent variables. Previous studies have extensively shown the effectiveness of the LDA for unsupervised learning of the latent representation of topics \cite{Cheng_2014}, social users \cite{Sun_2017}, and communities \cite{Xianghua_2013,Yonggan_Li_2016}, to name but a few. We are not, however, aware of work on latent aspect modeling except for that of a variational autoencoder by \cite{Fei_2021} where a review $r$ and its sentences $s_r$ were encoded to two sets of latent variables that follow Gaussian distribution: \textit{i}) coarse-grain review-level ${z_r}$, and \textit{ii}) fine-grain sentence-level $z_{s_r}$, followed by Transformer-based hierarchical classification for the task of review's sentiment classification. In their work, \citet{Fei_2021} assume that a review is formed by a bag of aspect and opinion words that are explicitly mentioned while filtering out other words using opinion lexicons. Successful for the task of sentiment classification, \citet{Fei_2021}'s unsupervised method detects explicit aspects and overlooks their latent occurrences. However, our main goal is to propose an unsupervised method to detect latent aspects when they are hidden with no surface form. 

\section{Proposed Method}
Let $\mathcal{D}$ be the set of all unique words in a language. Given a review $r_p = \{w_i\} \subseteq \mathcal{D}$ as a bag of its constituent words about a product or a service $p$ concerning an aspect $a \in A_p \subseteq \mathcal{D}$ which is not necessarily in $r_p$, that is, $a \notin r_p$, the latent aspect detection aims at identifying aspect $a$ in $r_p$. $A_p$ is the set of all aspects of the product or service $p$ and $R_p$ is the set of all $p$'s reviews. 

We propose to model aspects as latent variables. We employ latent Dirichlet allocation to learn the latent aspects distributions for generating the reviews. Specifically, we assume that a customer undergoes a two-stage hypothetical generative process to write a review about an aspect of a product or service: 1) \textit{deciding} on an aspect $a$ of the product or service amongst the set of aspects available for the product or service ${A_p}$ but with a different probability, and 2) \textit{selecting} the words from the set $\mathcal{D}$ of words in a language but with a different probability that is more interrelated to $a$ to express her opinion about $a$. In this generative process, a review is about all aspects of the product or service according to a probability distribution, i.e., $r_p \sim \theta_{r_p}$, and an aspect $a$ is not an explicit word but a latent probability distribution of all words with respect to $a$, i.e., $a \sim \varphi_a$, where ${\theta}$ and ${\varphi}$ follow Dirichlet probability distribution. Formally, the generative process for a set of reviews $R_p = \{r_p\}$ is:
\begin{enumerate}
    \item For each $r_p \in R_p$, draw $\theta_{r_p} \sim \text{Dir}(\bm{\alpha})$, where $\bm{\alpha}$ is a ${|A_p|}$-dimensional vector of positive reals which typically is sparse since a review is assumed to be about one or a few aspects of the product not all.
    \item For each $a \in A_p$, draw ${\varphi_a \sim  \text{Dir}(\bm{\beta})}$, where ${\bm{\beta}}$ is a ${|\mathcal{D}|}$-dimensional vector of positive reals. ${\bm{\beta}}$ is also assumed to be sparse since an aspect is interrelated to a few words and not to all words. 
    \item Given ${r_p \in R_p}$ along with its $\theta_{r_p}$:\\
 3.1. Draw an aspect ${a \sim \text{Multinomial}(\theta_{r_p})}$ \\
 3.2. For each of the words ${w_i}$ in ${r_p}$, draw  ${w_i \sim }$ $\text{Multinomial}({{\varphi}_a})$ \\ 
\end{enumerate}
Given the observed review set of a product ${R_p}$, our task is to infer $\theta_{r_p}$ for all reviews ${r_p \in R_p}$ and ${\varphi_a}$ for all aspects ${a \in A_p}$ of the product and estimation of the ${\bm{\alpha}}$ and ${\bm{\beta}}$ parameters using Gibbs sampling \cite{XiaoS_2010}\footnote{Our codebase and PxP dataset have been made publicly available at \href{https://github.com/MohammadForouhesh/latent-aspect-detection}{https://github.com/MohammadForouhesh/latent-aspect-detection}}.

\section{Experiment}
We show the empirical performance of our proposed method for $p=restaurant$. We trained our proposed model on datasets scraped from Google reviews of restaurants\textsuperscript{1} (PxP) across North America, and evaluate it on SemEval datasets \cite{Pontiki_2014,Pontiki_2015,Pontiki_2016} as the golden standards. Table ~\ref{stats} shows the detailed statistics of training and testing datasets.

\begin{table}[]
\begin{tabular}{lllll}
\cline{3-5}
                    &              & \multicolumn{3}{c}{\textbf{SemEval}}          \\ \hline
\textbf{Dataset}    & \textbf{PxP}\textsuperscript{1} & \textbf{2014} & \textbf{2015} & \textbf{2016} \\ \hline
{\#reviews}       & 71,842       & 2,068         & 1,075         & 1,393         \\
{\#unique words}  & 70,578       & 4,451         & 2,777         & 3,285         \\
{avg \#sentences} & 1.48         & 0.93          & 0.91          & 0.89          \\
{avg \#words}     & 20.36        & 16.85         & 14.99         & 14.9          \\ \hline
\end{tabular}
\end{table}
We removed numerical and non-English words as well as stop-words, emojis, and punctuations from reviews. We used flair's part-of-speech tagger \cite{flair_2019}, and lemmatized extracted phrases according to their part-of-speech.

\subsection{Evaluation Strategy}
\begin{figure}
\centering
\includegraphics[width=0.7\columnwidth]{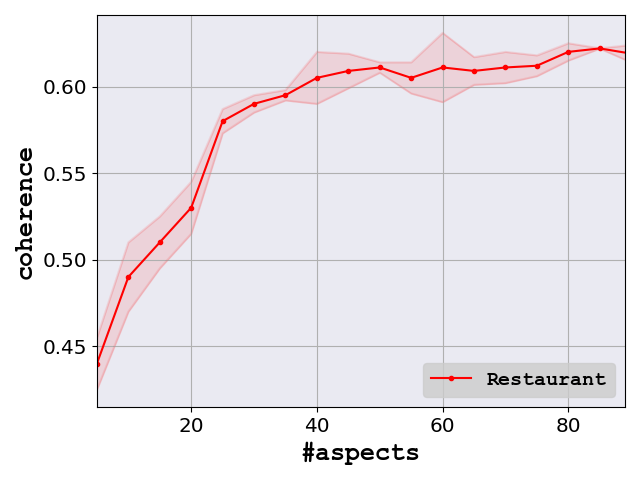}

\caption{Coherence score for PxP restaurant dataset.}
\label{fig:coherence}
\end{figure}

We estimate ${\theta}$ and ${\varphi}$ of our proposed model on Google reviews of restaurants with no supervision about whether the reviews have surface form for aspects. We trained the model for different numbers of aspects from 5 to 100 for restaurants and used the coherence score \cite{Mimno_2011} to find the optimum one. We set ${\bm{\alpha}}$ and ${\bm{\beta}}$ priors to 5.0 divided by the number of aspects and 0.01, respectively. As seen in Fig.~\ref{fig:coherence}, the coherence reaches a plateau at 30 aspects for restaurants.

To evaluate our trained model, we used SemEval datasets whose reviews have aspects with surface form (explicit) and are annotated. Given a review from SemEval, we removed the aspect from the review (make it latent) and used our model to infer the review's latent aspect as the one with the highest probability among all aspects of restaurants, i.e., ${a = max (\theta \sim \text{Dir}(\bm{\alpha}))}$. The predicted latent aspect $a$ is a probability distribution of words that should be compared to the removed aspect. A true prediction of latent aspect $a$ should have the surface form of the removed aspect in the top-${k}$ most probable words of ${a \in \varphi_a}$. 

For our training and test datasets are disjoint, we may face out of vocabulary (OOV) at evaluation. For instance, the word \textit{`sushi'} has been labeled as an aspect in SemEval reviews while our training dataset misses it. In this case, a true prediction of latent aspect $a$ should have the \textit{most semantically similar} words to the removed explicit aspect (e.g., \textit{`sushi'}) in the top-${k}$ most probable words of ${a \in \varphi_a}$ (e.g., \textit{`food'},\textit{ `dish'}). We used the Resnik similarity score \cite{Resnik_1999} to calculate the inter-word semantic similarities. 

We report mean reciprocal rank (MRR), recall,  nDCG,  and success (hit) at 5 as evaluation metrics for the performance of our trained model. 





\begin{figure*}[!ht]
\centering
\includegraphics[width=\textwidth]{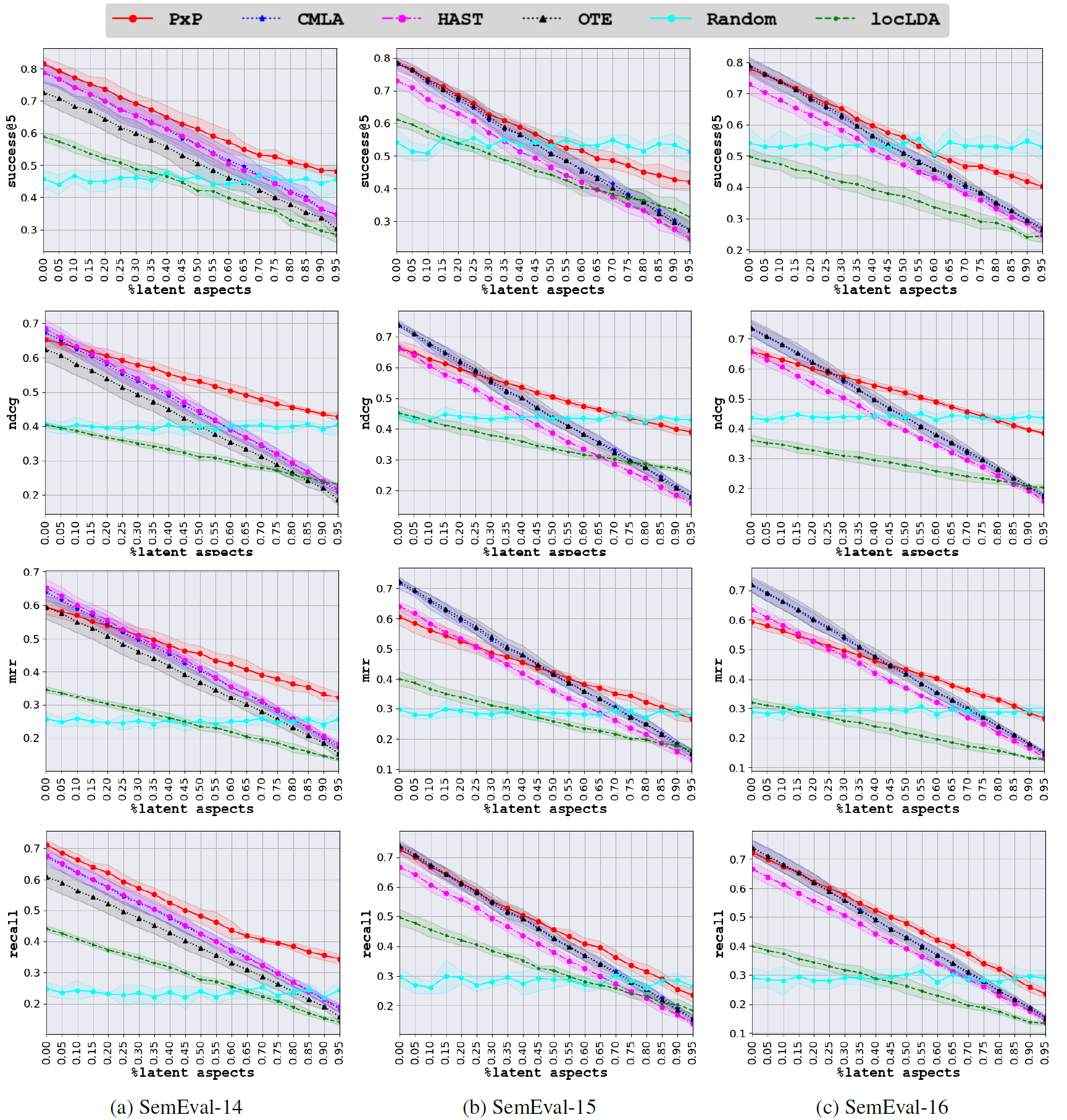}

\caption{Comparison results for the baselines and our model across different SemEval test sets.}
\label{fig:experimet_results}
\end{figure*}
\subsection{Baselines}
We benchmarked our proposed model against the following baselines:

\noindent\textbf{Random}: is a naive method that chooses a random noun word as the latent aspect.

\noindent\textbf{locLDA} \cite{Brody_2010} is an unsupervised method to identify explicit aspects. This method uses the standard implementation of LDA, but they failed to consider that a review may have no surface form for aspects.

\noindent\textbf{CMLA} \cite{Wang_2017} is a supervised aspect-opinion co-extraction model that leverages attention mechanisms. 

\noindent\textbf{HAST} \cite{LiBLLY18} is a supervised model on the tasks of explicit aspect detection. It takes into account the history of aspects with an attention block using bi-directional LSTMs.

\noindent\textbf{OTE-MTL} \cite{Zhang_2020} is a supervised multitask learning framework to extract aspect and opinion words jointly and parse sentiment dependencies between them. 

\noindent\textbf{PxP} is our proposed model. In contrast to supervised baselines, our model is trained without any direct human supervision. Also, unlike \textbf{locLDA}, it takes into consideration that a review might have a latent aspect with no surface form. 

\subsection{Results}
In this section, we seek to answer two research questions when dealing with noisy, short, unsolicited reviews in which aspects are latent:

\noindent\textbf{RQ1.} Does unsupervised modeling of aspects as latent variables lead to more accurate aspect detection compared to an unsupervised baseline that detect explicit aspects? 

\noindent\textbf{RQ2.} Do unsupervised methods fare better in the task of latent aspect extraction compared to supervised methods?

Fig.~\ref{fig:experimet_results} summarizes the comparative results for customer reviews for restaurants in terms of evaluation metrics. X-axis shows to what extent aspects are removed, as explained in the evaluation strategy. For example, at 0.3, we hide labeled explicit aspects in 30\% of SemEval's reviews and use the baselines to infer them. As seen, all of the baselines suffer from masking of explicit aspects in the restaurant domain but at different rates. In response to \textbf{RQ1}, it is evident that our unsupervised model consistently outperforms the state-of-the-art unsupervised baseline \textbf{(loc-LDA)} in terms of all evaluation metrics and across three different test sets. As shown, modeling aspects as latent variables helps with better detection of latent aspects in the context of noisy and short restaurant reviews. 

Regarding \textbf{RQ2}, our proposed unsupervised method fares better in the task of latent aspect extraction compared to the state-of-the-art supervised method \textbf{(CMLA)} as the number of latent aspects increases. For instance, on SemEval-14 dataset, our model \textbf{(PxP)} could consistently maintain higher MRR score when aspects are latent in more than 20\% of reviews. Our model predicts a ranked list of possible aspects for a given review, even for a review whose words were totally unseen in the training phase. The supervised baselines however fall short and their performance decreases linearly as the number of latent aspects increases. 

\section{Conclusion}
In this paper, we proposed an unsupervised LDA-based method to detect latent aspects of products and services in unsolicited reviews which are noisy, short and aspects are not explicitly mentioned due to social background knowledge. We assume that writing a review is a generative process of deciding on an aspect amongst the set of all available aspects for a product or service, and then writing the most interrelated words for the chosen aspect. In contrast to supervised methods, our model is trained without direct human supervision, and unlike existing unsupervised methods, it takes into consideration that a review might have a latent aspect. Benchmark results show that our model could outperform supervised and unsupervised baselines when reviews happen to have latent aspects.
\bibliography{anthology,custom}
\bibliographystyle{acl_natbib}




\end{document}